\documentclass[11pt]{article}

\usepackage[preprint]{acl}
\usepackage{booktabs}
\usepackage{times}
\usepackage{latexsym}
\usepackage{multirow}
\usepackage{amsmath}
\usepackage{comment}
\usepackage[table,dvipsnames]{xcolor}
\usepackage{enumitem}
\usepackage[T1]{fontenc}
\usepackage{colortbl}
\usepackage[utf8]{inputenc}

\usepackage{microtype}

\usepackage{inconsolata}
\usepackage{multirow}
\usepackage{makecell}
\usepackage{longtable}

\usepackage{graphicx}

\usepackage{xcolor}
\newcommand{\flg}[1]{\colorbox{red!20}{\textsc{#1}}}
\newcommand{\ufl}[1]{\colorbox{green!20}{\textsc{#1}}}
\newcommand{\gun}[1]{\colorbox{red!20}{\texttt{#1}}}
\newcommand{\gsf}[1]{\colorbox{green!20}{\texttt{#1}}}
\newcommand{\gco}[1]{\colorbox{orange!30}{\texttt{#1}}}
\newcommand{\annh}{\fcolorbox{red!60!black}{red!15}{\textit{\textbf{Annotators}}}}
\newcommand{\annn}{\fcolorbox{green!50!black}{green!15}{\textit{\textbf{Annotators}}}}
\newcommand{\cpct}[2]{\colorbox{#1}{$#2$}}

\usepackage[normalem]{ulem}

\definecolor{tier1}{RGB}{175,195,240}   
\definecolor{tier2}{RGB}{200,215,245}   
\definecolor{tier3}{RGB}{218,228,248}   
\definecolor{tier4}{RGB}{232,238,251}   
\definecolor{tier5}{RGB}{244,247,253}   

\newcommand{\ulHigh}[1]{\uline{#1}}
\newcommand{\ulMod}[1] {\uline{#1}}
\newcommand{\ulWeak}[1]{\uline{#1}}

\newcommand{\cphi}[2]{\colorbox{#1}{#2}}       
\newcommand{\cp}[2]  {\colorbox{#1}{#2}}       

\title{\emph{Epistemic Injustice in Language Models}:\\An Audit of Pretraining Filters and Guardrails}

\author{
 \textbf{Marco Antonio Stranisci\textsuperscript{1,2,*}},
 \textbf{A Pranav\textsuperscript{3,*}},
 \textbf{Rossana Damiano\textsuperscript{1}},
 \\
 \textbf{Christian Hardmeier\textsuperscript{2}},
  \textbf{Anne Lauscher\textsuperscript{3}}
\\
\\
 \textsuperscript{1}University of Turin, Italy
 \textsuperscript{2}IT University of Copenhagen, Denmark\\
\textsuperscript{3}Trustworthy AI Lab, University of Hamburg, Germany\\
  \small{
   \textbf{Correspondence:} \href{mailto:marcoantonio.stranisci@unito.it}{marcoantonio.stranisci@unito.it}, \href{mailto:pranav.agrawal@uni-hamburg.de }{pranav.agrawal@uni-hamburg.de } 
  }
}

\begin{document}
\maketitle
\def\thefootnote{*}\footnotetext{The first two authors contributed equally.}
\begin{abstract}
Modern language models rely on pretraining filters to remove undesirable content from training corpora and inference-time guardrails to suppress undesirable outputs during deployment. In this paper, we examine how these filtering and moderation decisions produce forms of epistemic erasure and reveal tensions both across automated systems and between these systems and human judgment. We audit four pretraining filters and three inference-time guardrails on Common Crawl sentences containing gender and regional-origin mentions, together with a manually annotated subset of 500 sentences. 
Our analysis shows that filtering and guardrail decisions are strongly associated with blocklist-based lexical cues, while frequently failing to flag content containing private information or explicit hate speech. At the same time, marginalized groups, particularly transgender people, women, and Central Americans, are significantly over-flagged across systems. Human annotators, by contrast, would retain 88.5\% of filter-flagged and 91.3\% of guardrail-flagged content, often recognizing representational  harms arising from tensions of content removal that current systems fail to capture. 
Taken together, our findings document a form of \textit{epistemic erasure} in which mentions of marginalized groups are disproportionately removed before pretraining and additionally suppressed again at inference time.\footnote{Code and Data are available: \url{https://anonymous.4open.science/r/Epistemic-Injustice-in-LLM-7EC0/README.md}}
\end{abstract}
 
\section{Introduction}

Designing safe Large Language Models (LLMs) remains a central challenge in today's AI research \citep{bender2021dangers}. However, reducing harm in LLM outputs is not merely a technical problem. Decisions about what content should be suppressed or permitted are inherently socio-technical and depend on norms, context, and the perspectives of diverse social groups \citep{blodgett2020language}. As a result, effective alignment requires sensitivity  across identities and communities, a challenge increasingly discussed under the framework of pluralistic alignment \citep{sorensen2024roadmappluralisticalignment}.
\begin{table}[t]
\centering
\setlength{\tabcolsep}{3pt}
\renewcommand{\arraystretch}{1.05}
\footnotesize
\begin{tabular}{@{}p{0.48\columnwidth} l l@{}}
\toprule
\textit{``Even though the early seventies was reveling in its first flush of feminism, a Black woman singing aggressively about sexual gratification went over like a pimp at a day care center.''}
& \begin{tabular}[t]{@{}l@{}}\ufl{Hatebase}\\ \flg{Shutterst.}\\ \ufl{DOLMA}\\ \ufl{HateBERT}\end{tabular}
& \begin{tabular}[t]{@{}l@{}}\gsf{Llama-Guard}\\ \gun{MD-Judge}\\ \gun{Qwen3Guard}\\ \annn\end{tabular} \\
\midrule
\textit{``The most compelling characters are a blonde baby dyke and the African-American drag queen, surely a stand-in for the late activist Sylvia Rivera.''}
& \begin{tabular}[t]{@{}l@{}}\ufl{Hatebase}\\ \flg{Shutterst.}\\ \ufl{DOLMA}\\ \ufl{HateBERT}\end{tabular}
& \begin{tabular}[t]{@{}l@{}}\gsf{Llama-Guard}\\ \gun{MD-Judge}\\ \gun{Qwen3Guard}\\ \annn\end{tabular} \\
\midrule
\textit{``Not saying hostile missile attacks can't be a lot of fun\ldots{} Cheerio LOL, no shit.''}
& \begin{tabular}[t]{@{}l@{}}\ufl{Hatebase}\\ \flg{Shutterst.}\\ \flg{DOLMA}\\ \flg{HateBERT}\end{tabular}
& \begin{tabular}[t]{@{}l@{}}\gun{Llama-Guard}\\ \gun{MD-Judge}\\ \gco{Qwen3Guard}\\ \annn\end{tabular} \\
\midrule
\textit{``Once you have downloaded the cracked version, you need to disable all security measures, such as antivirus and firewall protection.''}
& \begin{tabular}[t]{@{}l@{}}\ufl{Hatebase}\\ \ufl{Shutterst.}\\ \ufl{DOLMA}\\ \ufl{HateBERT}\end{tabular}
& \begin{tabular}[t]{@{}l@{}}\gun{Llama-Guard}\\ \gun{MD-Judge}\\ \gun{Qwen3Guard}\\ \annh\end{tabular} \\
\bottomrule
\end{tabular}
\caption{Common Crawl sentences with \textsc{filters}, \texttt{guardrails}, and pooled \textit{\textbf{annotator}} verdicts. \colorbox{green!20}{Green} = unflagged/safe/no harm, \colorbox{red!20}{red} = flagged/unsafe/harmful, \colorbox{orange!30}{orange} = controversial (Qwen only).}
\label{tab:examples_compact}
\end{table}
Two main approaches have emerged to mitigate harmful behavior in LLMs: First, data filtering strategies remove undesirable content from training corpora prior to pretraining \citep{albalak2024survey}. Second, inference-time guardrails detect harmful prompts or outputs and suppress unsafe generations \citep{inan2023llamaguardllmbasedinputoutput}. While these technologies are becoming standard components of modern LLM pipelines, little research has examined their impact on marginalized social groups \citep{dodge-etal-2021-documenting,luccioni2021s}. Although bias in the related area of hate speech detection has been studied~\cite[e.g.,][]{sap2019risk}, to date, no work has systematically assessed (\emph{i})~the broader representational effects of filtering and moderation systems used in modern LLM pipelines, or (\emph{ii})~the extent to which these systems align with one another and with human judgment. This gap is problematic because modern filtering and moderation systems shape the epistemic boundaries of LLMs used by millions of people worldwide, but their social consequences remain poorly understood.


In this work, we pose the following research question: \emph{\textbf{Whose identity mentions are disproportionately removed by pretraining filters and inference-time guardrails, and are the observed patterns consistent across systems and with human judgment?}}
To answer this question, we propose a framework for evaluating filtering strategies and guardrails on harm detection, measuring their convergence, identifying their respective impact on vulnerable identities, and assessing their alignment with human judgments. 
We test the framework on four data filtering strategies and three guardrails using a sample of documents drawn from Common Crawl.
Here, we focus on identity mentions related to gender and origin: categories that are central to the bias literature and well-suited to extraction via curated lexicons derived from Wikidata \citep{erxleben2014introducing} and Named Entity Recognition (NER). Table~\ref{tab:examples_compact} gives a snapshot of our results.
The first two rows show \textit{epistemic erasure}, where systems flag mentions of Black women and queer identities for removal while annotators would keep them.
The next two rows show keep-or-remove \textit{tensions}.
Our experiments on a sample of Common Crawl sentences yield three main findings:

\paragraph{F1. \textit{Disagreement} of filters and guardrails.}
We find that guardrails correlated with blocklist based lexicons, and filters miss most of what their own taxonomies define as harmful, including privacy violations, intellectual property, and weapons.

\paragraph{F2. \textit{Epistemic erasure} of marginalised identities.} 
Our findings show that mentions of marginalized communities, particularly transgender people, women, and Central Americans, are significantly over-flagged by guardrails and filters, despite having
been trained on different taxonomies by organisations on different continents.

\paragraph{F3. \textit{Tensions} in content moderation.} Our results show that humans disagree with automated systems at very high rates, they would not
remove content flagged by a filtering strategy in 88.5\%, and content flagged by a
guardrail in 91.3\% of cases.

In sum, our findings suggest that filtering and moderation systems do not merely reduce harmful content, but actively shape whose identities, perspectives, and experiences remain visible in modern language models.

\section{Related Work}
\label{sec:related}

\paragraph{Data Filtering Strategies.}

Filters at pretraining time strip harmful or low-quality text from web corpora before models train on them \citep{penedo2024finewebdatasetsdecantingweb, gao2020pile800gbdatasetdiverse}.
Toxicity filtering reduces unsafe generation at the cost of benchmark performance \citep{longpre-etal-2024-pretrainers}, and audits of web-scale corpora document harmful content that current filters do not catch \citep{luccioni2021s, mendu2025saferpretraininganalyzingfiltering}.
At inference time, guardrail models classify inputs and outputs against an explicit harm taxonomy supplied in the prompt \citep{inan2023llamaguardllmbasedinputoutput, ghosh2024aegisonlineadaptiveai, 10.1609/aaai.v37i12.26752}.
Their agreement with human raters is moderate, even on examples where annotators agree with each other \citep{li-etal-2024-salad, ji2023beavertailsimprovedsafetyalignment}.

\paragraph{Bias in automated harm detection.}
Extensive literature shows that Hate speech classifiers reproduce the stereotypes their annotators hold \citep{sap2019risk,davani2021hatespeechclassifierslearn}.
These biases connect with issues in data and annotation practices \citep{vidgen2020directions}, which leave downstream LLMs under-representing non-Western regions and homogenizing cultural representation toward Western norms \citep{schwobel-etal-2023-geographical, qadri-etal-2025-risks}.
Harm-detection systems extend the pattern to gender-queer dialect, misclassifying reclaimed slurs that queer speakers use as in-group identifiers \citep{draetta2024reclaim} and performing worst on texts authored by the very communities they are meant to protect \citep{dorn-etal-2024-harmful}.

\section{Methodology}
\label{sec:method}

We compare four pretraining filters and three guardrail models on a sample of Common Crawl sentences.
The corpus, the systems, the analytical splits, the annotation procedure, the inductive coding of the annotators' justifications, and the statistical conventions are described below.

\subsection{Data Collection}
\label{sec:data}

The source dataset of our experiment is a sample of documents gathered from a Common Crawl (CC) snapshot.\footnote{\url{https://data.commoncrawl.org/crawl-data/CC-MAIN-2024-33/index.html}}
We applied a language detection tool\footnote{\url{https://pypi.org/project/langdetect/}} to retain only English texts and applied quality-filtering heuristics \citep{raffel2020exploring} to remove documents with fewer than $5$ sentences and fewer than $10$ words per sentence on average.
This approach yielded a dataset of $102{,}727$ documents.

\subsection{Identity Extraction}
\label{sec:identities}

Next, we performed an identification of identities related to gender and origin in documents at the sentence-level in two steps: extraction and linking.
We extracted all the entities of the type PERSON with a tool for Named Entity Recognition (NER).\footnote{\url{https://spacy.io/}}
We extracted demonyms and gender mentions through string pattern matching, using as a reference Wikidata entities of the type \textit{gender identity} (Q48264) and \textit{country} (Q6256).
We linked all the entities of the type PERSON to their corresponding Wikidata ids through Wikidata APIs\footnote{\url{https://www.mediawiki.org/wiki/API:Search}} and gathered their gender and country of birth.
We then mapped identity mentions to macro-categories through manually curated resources, all aligned with Wikidata entities:

\begin{itemize}[nolistsep]
    \item Following gender taxonomy proposed by \citet{queer-in-ai-dni-guide}, we encode gender as five categories (man, woman, non-binary/genderqueer/third gender, questioning, genderfluid/gender non-conforming) and treat modality (transgender vs.\ cisgender) as an orthogonal axis.
    \item A lexicon of the seventeen United Nations Macro-Regions \citep{united1982standard} mapped against all the instances of \textit{country} in Wikidata.
\end{itemize}

After dropping ambiguous or low-frequency mentions, the trans/cis lexicon retained $47{,}031$ sentences, the gender lexicon retained $70{,}925$ sentences across the five categories, and the world-region lexicon retained $9{,}975$ sentences across seventeen regions.
Of the seventeen regions, ten reach the inferential threshold of $n\geq 50$ sentences and enter the analysis; the remaining seven appear in the Appendix.

\subsection{Analytical Splits}
\label{sec:splits}

Two splits structure the analysis to disentangle the effects of filters and guardrails.
The \emph{flagged set} is the $72{,}659$ sentences flagged by at least one filter; it carries the inter-system alignment analysis and the identity-conditional flag-rate analysis.
The \emph{unflagged set} is the $257{,}210$ sentences sampled from the same $72{,}659$ source documents on which no filter fired; it carries the guardrail-only analysis in Table~\ref{tab:blindspot_top3}.
Because the two splits are drawn from the same document pool, a guardrail verdict on a sentence in the unflagged set is directly comparable with a guardrail verdict on a sentence in the flagged set from the same document.

\subsection{Harm Detection Systems Selection}
\label{sec:filters}

\paragraph{Filtering Strategies.}
We selected four filters that have been extensively documented and adopted in previous literature (see \citet{Stranisci_2026}) and that belong to two approaches: rule-based lexicon and classifier-based detectors.

\textsc{Hatebase} is the Davidson-refined slur lexicon, focused on racial and misogynist terms \citep{davidson2017automatedhatespeechdetection}.
\textsc{Shutterstock} is the sexual-vocabulary lexicon used in the C4 and T5 pretraining filters \citep{raffel2020exploring, dodge-etal-2021-documenting}.
\textsc{DOLMA} is the FastText classifier trained on Jigsaw and shipped with the \textsc{DOLMA} pretraining recipe \citep{soldaini-etal-2024-dolma}.
\textsc{HateBERT} is BERT re-trained on RAL-E, a corpus of posts from Reddit communities banned for abusive content \citep{caselli-etal-2021-hatebert}.
The four filters cover the three published filter families and span the two harm regions that pretraining filters most often target: hateful or toxic discourse (\textsc{Hatebase}, \textsc{HateBERT}) and explicit sexual content (\textsc{Shutterstock}), with \textsc{DOLMA} as a general-purpose toxicity classifier \citep{Stranisci_2026}.

\textsc{Hatebase} and \textsc{Shutterstock} return binary verdicts.
\textsc{DOLMA} and \textsc{HateBERT} return continuous scores; we apply each model's released default threshold to obtain a binary verdict and clip \textsc{DOLMA}'s raw logits to the unit interval for the confidence-based analyses.

\paragraph{Guardrail Models.}
We selected three guardrail models of comparable size (8B parameters) released by organisations on three different continents, so that any divergence between them cannot be attributed to scale alone and can plausibly be related to the harm taxonomies their developers encode.

\texttt{Llama-Guard-3-8B} is the third-generation Meta guardrail, fine-tuned on a flat hierarchy of fourteen risk categories derived from Meta's usage policy \citep{inan2023llamaguardllmbasedinputoutput}.
\texttt{MD-Judge-v0.1} is the LLM-based safety judge introduced with SALAD-Bench, fine-tuned on Mistral-7B over a three-level taxonomy spanning six domains and 66 categories \citep{li-etal-2024-salad}.
\texttt{Qwen3Guard-Gen-8B} is the eight-billion-parameter generative variant of Alibaba's \texttt{Qwen3Guard} family, trained over a multilingual harm taxonomy that introduces a third \textit{controversial} label for cases whose safety status depends on context or policy.\footnote{We use the generative variant with chain-of-thought disabled (\texttt{enable\_thinking=False}). The released model card describes the taxonomy and training procedure.}

Each guardrail receives a sentence as user input and returns a binary safe/unsafe label (or a three-way label for \texttt{Qwen3Guard}) together with one or more harm categories.
\texttt{MD-Judge} returns its categories as integer codes mapped to the SALAD-Bench taxonomy.
For the binary analyses we fold \texttt{Qwen3Guard}'s \textit{controversial} responses into \textit{unsafe}, and we report sensitivity rows that retain the three-way labelling scheme.

\subsection{Manual Annotation}
\label{sec:annotation}

We selected a subset of $500$ sentences and annotated them for harm.
Two undergraduate students in Linguistics, a man and a woman, who joined the research team during their internship, annotated the subset.
Sentences were sampled with stratification across filter--guardrail agreement patterns to oversample disagreement cases.

\paragraph{Annotation Task.}
Given the well-documented subjectivity inherent in toxicity-annotated corpora \citep{davani2024disentangling,lee2024exploring}, we framed each decision as a tradeoff between the cost of retaining a flagged document and the cost of removing it, rather than as a direct judgement of toxicity.
Annotators chose whether to remove a document as a binary judgement and motivated, where they wished to, the potential harm introduced by keeping or removing the document through open-ended questions: \textbf{[T1]} Would you remove this document? [yes-no];\textbf{[T2]} Which harm can happen if the document is kept?; \textbf{[T3]} Which harm can happen if the document is removed?

\paragraph{Annotation Procedure.} Each sentence received $2$ independent annotations on all questions.
Inter-annotator agreement on T1, reported as Cohen's $\kappa$, is $0.46$ with $85\%$ of annotations in agreement.
Annotators then provided justifications for T1 and T2 in a free-text form. 

The free-text justifications from T2 and T3 were then inductively coded.
Both authors independently developed candidate category sets through open coding on the full set of justifications; the first author then applied the codes across all justifications.
The authors converged on the final categories through discussion, yielding seven harms-of-keeping categories and seven harms-of-removing categories (Table~\ref{tab:harms}).

\subsection{Evaluation Setting}
\label{sec:stats}

\paragraph{Agreement audit on filters and guardrails.}
We treat each filter and guardrail as a binary indicator on the sentence and measure pairwise agreement with Cohen's $\kappa$, with $95\%$ confidence intervals from $1{,}000$ cluster-bootstrap iterations that resample at the document level.
The resulting $7\times 7$ matrix places filters and guardrails as shown in Figure \ref{fig:correlation}.
We report $\kappa$ on the combined set of $308{,}308$ sentences and on the flagged subset of $72{,}659$ sentences (those flagged by at least one filter).
On the unflagged subset of $257{,}210$ sentences (those no filter flagged) we report each guardrail's top unsafe categories as a share of that guardrail's unsafe verdicts on the subset, which is shown in Table \ref{tab:blindspot_top3}.

\paragraph{Identity-conditional flag rates.}
For each filter and guardrail we report the flag rate on each identity group with Wilson $95\%$ confidence intervals.
Effect sizes are rate ratios against a pooled reference: cisgender for the trans/cis modality, man for the five gender categories, and the pooled non-target regions for each world region.

\paragraph{Significance and display conventions.}
Within each identity dimension we adjust $p$-values with the Benjamini--Hochberg procedure (BH-FDR) and report them as $q$.
Statistical values appear in tiered pastel-blue colourboxes that encode magnitude or significance, with darker shades marking larger $|\kappa|$ values, or smaller $q$ values: \cphi{tier1}{$\kappa=0.60$}, \cphi{tier2}{$\kappa=0.45$}, \cphi{tier3}{$\kappa=0.25$}, \cphi{tier4}{$\kappa=0.15$}, \cphi{tier5}{$\kappa=0.05$}; \cp{tier1}{$q<.001$}, \cp{tier2}{$q<.01$}, \cp{tier3}{$q<.05$}, \cp{tier4}{$q<.10$}, \cp{tier5}{$\textrm{ns}$}.
Figure markers follow the same tiering: $\bullet$ marks $q<.001$, $\circ$ marks $q<.05$, and no marker indicates $q\geq.05$.
Verbal descriptors of statistical strength are typeset in italic with a solid underline, e.g., \ulHigh{\emph{strong}}, \ulMod{\emph{significant}}, \ulWeak{\emph{not significant}}, \ulWeak{\emph{uncorrelated}}.

\section{Findings}

In this section, we show our findings on agreement audits between filters and guardrails, show the account of epistemic erasure of the identities and give a walkthrough of tensions in content moderation with examples. 

\subsection{Study of \emph{disagreement} of filters and guardrails}
\label{sec:rq1}

\begin{figure}[t]
    \centering
    \includegraphics[width=\linewidth]{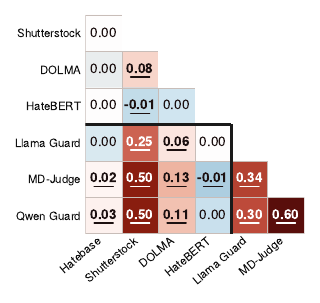}
    \caption{Pairwise Cohen's $\kappa$ between filter and guardrails. Bold marks $q<.05$; underline marks $q<.01$.}
    \label{fig:correlation}
\end{figure}

In this section we show how the four filters and three guardrails correlate (Figure~\ref{fig:correlation}) and which categories filters miss that guardrails catch (Table~\ref{tab:blindspot_top3}).

\paragraph{Guardrails are \ulHigh{\emph{correlated}} with the lexicon-based blocklist.}
\textsc{Shutterstock} is \ulWeak{\emph{significantly correlated}} with each guardrail at \cphi{tier3}{$\kappa_{\text{Sh,LG}} = 0.249$}, \cphi{tier2}{$\kappa_{\text{Sh,MD}} = 0.498$}, and \cphi{tier2}{$\kappa_{\text{Sh,Qw}} = 0.504$} (\cp{tier1}{$q<.001$} each). 
On the other hand,
\textsc{Hatebase}, \textsc{DOLMA}, and \textsc{HateBERT} are \ulWeak{\emph{weakly correlated}} with every guardrail.
\textsc{Shutterstock} is a blocklist: a sentence containing any of its sexual-vocabulary terms is removed, with no attention to context.
\citet{dodge-etal-2021-documenting} document that blocklists of this kind suppress non-pornographic content from queer folks, sex educators, and sex workers, alongside benign mentions of body parts and identity terms.
 
\paragraph{\texttt{Llama-Guard} is an outlier among the guardrails.}
\texttt{MD-Judge} and \texttt{Qwen3Guard} reach \ulHigh{\emph{substantial agreement}}: \cphi{tier1}{$\kappa = 0.604$} on the full set, rising to \cphi{tier1}{$\kappa = 0.651$} on the flagged set.
\texttt{Llama-Guard} reaches only \ulMod{\emph{moderate agreement}} with each: \cphi{tier2}{$\kappa = 0.340$} with \texttt{MD-Judge} and \cphi{tier3}{$\kappa = 0.301$} with \texttt{Qwen3Guard}, roughly half the agreement between \texttt{MD-Judge} and \texttt{Qwen3Guard}.
The three guardrails fall in similar bands against the four filters; what splits them is each other, with \texttt{MD-Judge} and \texttt{Qwen3Guard} clustering together and \texttt{Llama-Guard} apart.
\texttt{MD-Judge} was fine-tuned on the SALAD-Bench taxonomy of six domains and 66 categories \citep{li-etal-2024-salad}, and \texttt{Qwen3Guard} on a multilingual harm taxonomy in the same academic-research tradition.
\texttt{Llama-Guard}'s fourteen categories instead derive from Meta's usage policy \citep{inan2023llamaguardllmbasedinputoutput}, with heavy weight on intellectual property and privacy.
 
\begin{table}[t]
  \centering
  \small
  \renewcommand{\arraystretch}{1.15}
  \begin{tabular}{@{}llr@{}}
    \toprule
    \textbf{Guardrail} & \textbf{Category} & \textbf{\%} \\
    \midrule
    \multirow{3}{*}{\texttt{Llama-Guard}}
        & Intellectual Property      & \cpct{tier3}{34.7} \\
        & Privacy                    & \cpct{tier3}{33.9} \\
        & Indiscriminate Weapons     & \cpct{tier5}{6.7}  \\
    \midrule
    \multirow{3}{*}{\texttt{MD-Judge}}
        & Representation \& Toxicity & \cpct{tier1}{75.1} \\
        & Socioeconomic Harms        & \cpct{tier4}{11.8} \\
        & Information \& Safety      & \cpct{tier5}{9.9}  \\
    \midrule
    \multirow{3}{*}{\texttt{Qwen3Guard}}
        & Non-violent Illegal Acts   & \cpct{tier3}{24.0} \\
        & Politically Sensitive      & \cpct{tier3}{22.8} \\
        & Sexual Content             & \cpct{tier4}{19.7} \\
    \bottomrule
  \end{tabular}
  \caption{Top three unsafe categories per guardrail across sentences that no filter flagged, based on guardrail's unsafe verdicts. (Highlighted based on $q$ values)}
\label{tab:blindspot_top3}
\end{table}
 
\paragraph{Filters miss harm in intellectual property, privacy, and political content.}
Across the $257{,}210$ sentences no filter flagged, the three guardrails issued between $3{,}921$ and $10{,}800$ unsafe verdicts each (Table~\ref{tab:blindspot_top3}).
\texttt{Llama-Guard}'s verdicts on this subset are $68.6\%$ Intellectual Property ($34.7\%$) and Privacy ($33.9\%$) (\cp{tier1}{$q<.001$} for each), categories no filter in the audit is built to detect.
\texttt{Qwen3Guard}'s top four are Non-violent Illegal Acts ($24.0\%$), Politically Sensitive Topics ($22.8\%$), Sexual Content ($19.7\%$), and Unethical Acts ($17.1\%$) (all \cp{tier2}{$q<.01$}); only sexual content falls inside the filter family's coverage.
\texttt{MD-Judge}'s verdicts are $75.1\%$ Representation and Toxicity (\cp{tier1}{$q<.001$}), the one domain \textsc{Hatebase}, \textsc{HateBERT}, and \textsc{DOLMA} also target, where $2{,}948$ sentences pass every filter.
The four filters miss the categories guardrails flag most (intellectual property, privacy, political speech), and within representation and toxicity itself $2{,}948$ sentences go unflagged by all four.

\subsection{Study of \emph{epistemic erasure} of marginalised identities}
 
In this section, we report which identity groups filters and guardrails over-flag and under-flag.
Rates are compared against pooled-other baselines, with Wilson 95\% CIs and BH-FDR adjusted significance markers.
 
\begin{figure}[t]
  \centering
  \includegraphics[width=\columnwidth]{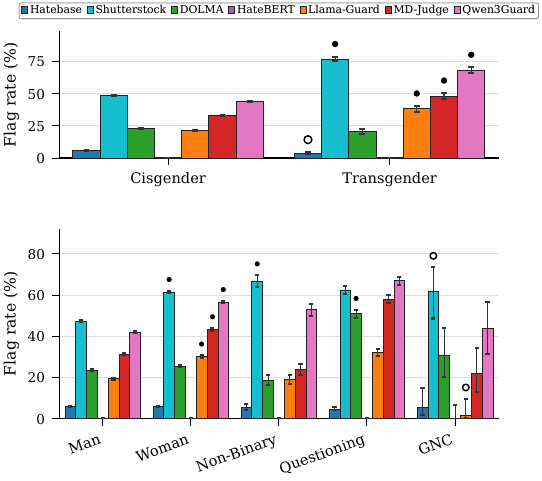}
  \caption{Flag rates by gender modality (top) and gender category (bottom), with Wilson 95\% CIs. Filled circles: \cp{tier1}{$q<.001$}; open circles: \cp{tier3}{$q<.05$} (BH-FDR).}
  \label{fig:bars_trans_gender}
\end{figure}
 
\paragraph{Central American mentions are flagged at \ulHigh{\emph{highly significant}} rates.}
\texttt{Llama-Guard} flags $95.9\%$, \texttt{Qwen3Guard} $98.7\%$, and \textsc{Shutterstock} $99.3\%$ (each \cp{tier1}{$q<.001$}); \texttt{MD-Judge} reaches $65.9\%$ (\cp{tier3}{$q<.05$}; Table~\ref{tab:app_origin}).
The \texttt{Llama-Guard} rate exceeds its $26.2\%$ pooled-other baseline by $69.7$ percentage points, the largest single deviation in the audit.
\textsc{DOLMA} stays at $8.1\%$ and \textsc{HateBERT} at $0.4\%$ on the same passages.
Flagging Central American content upstream could lead to geographical biases observed in language generation \citep{schwobel-etal-2023-geographical} and the \textit{cultural homogenization} documented across generative models \citep{qadri-etal-2025-risks}.
 
\paragraph{Transgender mentions are flagged \ulHigh{\emph{significantly}} more often than cisgender mentions.}
Flag rates rise from $48.5\%$ to $76.8\%$ for \textsc{Shutterstock}, $21.5\%$ to $38.1\%$ for \texttt{Llama-Guard}, $33.0\%$ to $48.1\%$ for \texttt{MD-Judge}, and $43.6\%$ to $68.3\%$ for \texttt{Qwen3Guard} (each \cp{tier1}{$q<.001$}; Table~\ref{tab:app_trans}), a ratio of $1.5$ to $1.8$ across the four systems.
\textsc{Hatebase} moves the other way ($5.7\%$ to $3.5\%$, \cp{tier3}{$q=.012$}) because its slur list does not include trans-coded vocabulary.
This epistemic erasure of transgender mentions has been linked to \textit{gender-non-affirmative language} and misclassification of reclaimed slurs from gender-queer authors in aligned models \citep{ovalle-etal-2025-root, dorn-etal-2024-harmful}.

\begin{figure}[t]
  \centering
  \includegraphics[width=\columnwidth]{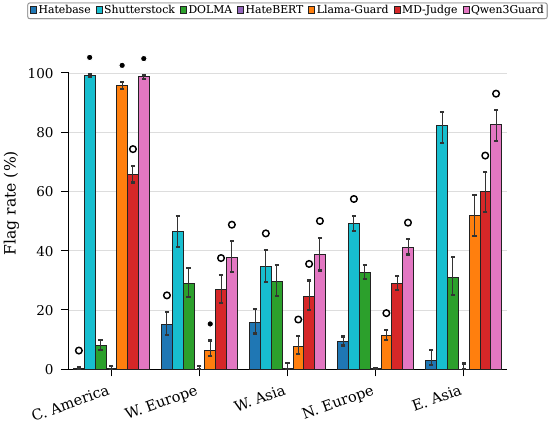}
  \caption{Flag rates for the five world regions with the largest departure from the overall mean baseline. Markers as in Figure~\ref{fig:bars_trans_gender}.}
  \label{fig:bars_origin_top5}
\end{figure}
 
\paragraph{Women's mentions are flagged \ulHigh{\emph{significantly}} more often than men's.}
\texttt{Llama-Guard} rises from $19.4\%$ on man to $30.1\%$ on woman, \texttt{MD-Judge} from $31.3\%$ to $43.3\%$, \texttt{Qwen3Guard} from $42.1\%$ to $56.5\%$, and \textsc{Shutterstock} from $47.3\%$ to $61.4\%$ (each \cp{tier1}{$q<.001$}; Table~\ref{tab:app_gender}).
Non-binary and questioning groups follow the same direction on smaller subsamples; only \textsc{Shutterstock} reaches \cp{tier1}{$q<.001$} on non-binary ($66.9\%$), and only \textsc{DOLMA} on questioning ($50.9\%$).
High flags of women's mentions has been linked to gender stereotyping in downstream generation \citep{sheng-etal-2019-woman, kotek-etal-2023-gender} and to \textit{representational harm} in how language models later depict women \citep{blodgett2020language}.
 
\paragraph{All three guardrails \ulHigh{\emph{significantly}} under-flag Western European, Western Asian, and Northern European mentions.}
\texttt{Llama-Guard} flags these three regions at $6.5\%$, $7.5\%$, and $11.5\%$ (\cp{tier1}{$q<.001$} for Western Europe, \cp{tier3}{$q<.05$} for the other two), rate ratios of $0.25$, $0.29$, and $0.44$ against its $26.2\%$ baseline.
\texttt{MD-Judge} flags Western Europe at $26.9\%$ and Western Asia at $24.7\%$ (both \cp{tier3}{$q<.05$}), each at roughly two-thirds of its baseline.
\texttt{Qwen3Guard} flags all three at $37.9\%$, $38.7\%$, and $41.3\%$ (all \cp{tier3}{$q<.05$}), each below its baseline.
Contrary to expectations, \texttt{Qwen3Guard} over-flags Eastern Asian mentions at $82.7\%$ (\cp{tier3}{$q<.05$}), above its baseline.
The convergence across guardrails points to \textit{homogenization} rather than to developer-region effects.

\begin{figure}[t]
    \centering
    \includegraphics[width=\linewidth]{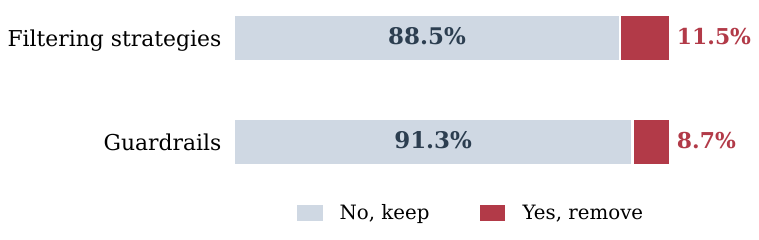}
    \caption{Annotators disagree with most system verdicts, retaining $88.5\%$ of filter-flagged and $91.3\%$ of guardrail-flagged content on the $500$-sentence subset. Bars show the per-system retention rate.}
    \label{fig:false_positive}
\end{figure}

\subsection{Study of \emph{tensions} in content moderation.}
\label{sec:rq3}
 
We compared annotator verdicts against the labels from the four filters and three guardrails on the $500$-sentence manual subset.
\textbf{Annotators disagree with system verdicts in the majority of cases.}
They would retain $88.5\%$ of filter-flagged content and $91.3\%$ of guardrail-flagged content (Figure~\ref{fig:false_positive}).
Based on inductive coding, the categories capture harms a sentence carries if kept, and harms incurred if it is removed (Table~\ref{tab:harms}).
We discuss the three largest categories on each side, alternating keep and remove pairs by content domain here.
 
\paragraph{Harm of \emph{keeping}: sexualised commodification.}
$41$ sentences were marked for removal as sexualised commodification of sex workers; one such sentence:
\begin{quote}\itshape
``Camming with toys has never been more thrilling, as camgirls explore the endless possibilities of interactive pleasure.''
\end{quote}
All four filters pass the sentence, but all three guardrails return unsafe.
One annotator justified the keep as \textit{``objectification and sexualization of sex workers,''} what \citet{mcglynn2025epistemic} terms epistemic fungibility, the treatment of women in sex work as interchangeable.
Pretraining on framings which has sexualised commodification could propagate gender bias that associates women with sex work \citep{sheng-etal-2019-woman}.
 
\paragraph{Harm of \emph{removing}: sex-worker visibility.}
Annotators recommended keeping $22$ sentences for sex-worker visibility, including:
\begin{quote}\itshape
``Our call girls in Hyderabad are professional, attractive and provide exceptional service.''
\end{quote}
The filters don't flag this, but all three guardrails return unsafe.
One annotator noted that \textit{``some sex worker could feel unrepresented''} if it were removed.
Above, this kind of content was marked for removal under sexualised commodification; here, it is marked for retention under occupational visibility.
Removing material like this from pretraining strips sex workers' occupational visibility from training corpora; \citet{xu2021detoxifying} show that similar identity-marker filtering can degrade model performance on the affected community.

\begin{table}[t]
\centering
\small
\setlength{\tabcolsep}{4pt}
\renewcommand{\arraystretch}{1.15}
\begin{tabular}{@{}l p{0.30\linewidth} r p{0.28\linewidth} r@{}}
\toprule
\textbf{Domain} & \textbf{Reason to \emph{remove}} & $n$ & \textbf{Reason to \emph{keep}} & $n$ \\
\midrule
Sex work        & Sexualised commodification          & $41$ & Sex-worker visibility            & $22$ \\
Race            & Racialised objectification          & $32$ & Group representation             & $37$ \\
Politics        & Politically slanted framing         & $17$ & Political viewpoint              & $54$ \\
Risk            & Soft invitations to risky behaviour & $35$ & Practical information            & $76$ \\
Hate            & Hate speech \& discrimination       & $14$ & First-person testimony           & $16$ \\
\bottomrule
\end{tabular}
\caption{Inductive codes on justifications of tensions of content moderation.}
\label{tab:harms}
\end{table}
 
\paragraph{Harm of \emph{keeping}: invitations to substance abuse.}
$35$ sentences were marked as invitations to substance abuse, such as:
\begin{quote}\itshape
``Sour Suckers THC Gummies For Sale\ldots\ Buy Flav Gummies Apple Rings 100mg THC Gummies For Sale Buy Faded Cannabis Co.''
\end{quote}
The filters don't flag it, \gun{Llama-Guard} and \gun{MD-Judge} call it unsafe, while \gco{Qwen3Guard} marks it as controversial under Non-violent Illegal Acts.
One annotator described the sentence as a \textit{``possible soft invitation to THC consume,''} the kind of drug-marketing copy that \citet{gomes2024problematizing} show platforms retain while removing harm-reduction speech.
Retained in pretraining corpora, models tend to reproduce brand names and purchase phrases in drug-related responses \citep{carlini2021extracting}.
 
\paragraph{Harm of \emph{removing}: practical how-to information.}
Annotators marked $76$ sentences for retention as practical how-to information; one such sentence:
\begin{quote}\itshape
``They usually scrape off a tiny bit of the rough skin to make tiny blood clots visible and examine the inside of warts.''
\end{quote}
All filters pass the sentence except \flg{DOLMA}, which flags it; \gsf{Llama-Guard} and \gsf{MD-Judge} return safe, while \gco{Qwen3Guard} marks it as controversial under Unethical Acts.
One annotator would retain the sentence because removal would mean a \textit{``loss of health related infos for people with the condition.''}
Filtering medical content of this kind from pretraining would reduce models' capabilities in the medical domains \citep{rathi2026shaping}.
 
\paragraph{Harm of \emph{keeping}: racialised objectification.}
$32$ sentences were marked for removal as racialised objectification, including:
\begin{quote}\itshape
``I am a young woman of high level for a courteous, respectful man to whom I offer a very exclusive service.''
\end{quote}
All four filters retain the sentence, while \gun{Llama-Guard} and \gun{MD-Judge} return unsafe and \gco{Qwen3Guard} marks it as controversial under Sexual Content.
One annotator glossed the sentence as \textit{``aiding prostitution and sexualization on a racial base,''} a representational harm that \citet{katzman2023taxonomizing} categorises as trafficking in stereotypes about women of colour.
Keeping sentences could lead to intersectional bias in models that links women of colour to sex work \citep{kirk2021bias}.
 
\paragraph{Harm of \emph{removing}: political viewpoints.}
$54$ sentences were marked for retention as political viewpoints, such as:
\begin{quote}\itshape
``In order to further the Democrat party's gargantuan voter drive as they allow massive numbers of foreigners into the country, they are using the guise of asylum to usher in as many\ldots''
\end{quote}
No filter flags the sentence; all guardrails flag it as unsafe.
One annotator flagged the removal as \textit{``loss of some political points of view.''}
Without such viewpoints in pretraining, models would lose the ability to represent them, propagating specific political leanings in model outputs \citep{feng2023pretraining}.

\section{Towards Epistemic Justice}
\label{sec:towards-epistemic-justice}
Our findings document a form of \textit{epistemic erasure} of marginalised identities at the pretraining stage and again at inference.
Addressing this requires socio-technical mitigations across two directions: making filters and guardrails more epistemically just, and using them to promote epistemic justice more broadly.

\paragraph{Epistemic justice for LMs.}
Filter lexicons should be released alongside documentation of how often they flag content involving different identity groups \citep{lucy-etal-2024-aboutme}.
Guardrail fine-tuning should involve reflexivity about the harm taxonomies and annotator pools that produce its training labels, rather than imposing a `view from nowhere' on what counts as harmful identity language.
Pretraining data collection and ownership should follow data sovereignty principles \citep{kukutai2016indigenous}, addressing the same biases at their source rather than only at the audit stage.

\paragraph{LMs for epistemic justice.}
Members of affected groups should be consulted about how their identity is represented in pretraining corpora \citep{jo2020lessons}, since defining what counts as identity-related erasure cannot be done by machine-learning methods in isolation from human judgement.
Filters and guardrails can then be red-teamed against that material to measure their over-flagging rates \citep{ganguli2022redteaminglanguagemodels}, and embedding retrieval or trained classifiers can locate further instances in pretraining corpora \citep{khandelwal2020generalizationmemorizationnearestneighbor}.

\paragraph{Conclusions.}
In this paper we presented an audit of pretraining filters and inference-time guardrails in the detection of harmful content in datasets that are widely used for pretraining.
All filters and guardrails over-flag marginalized groups, particularly, transgender mentions, women's mentions, and Central American content, across organisations on three continents and different harm taxonomies, showing that specific identities are systematically erased by these systems.
These erasure patterns are even more concerning when human evaluators are asked to assess the outcomes of filters and guardrails: evaluators would not remove $88.5\%$ of the content that filters flag and $91.3\%$ of the content that guardrails flag, observing that content removal is not only unnecessary but also a source of bias against specific categories of people.
Together, the findings show that filter and guardrail systems systematically erase content about marginalized identities, and that human evaluators would keep most of what these systems remove.
Future work will extend the audit to identity dimensions less amenable to lexicon-based extraction (socioeconomic status, religion, political belief), develop a dedicated annotation scheme with a more demographically diverse pool of annotators, and extend the framework to non-English corpora.

\section*{Limitations}
Current limitations of our research are the following: 
Only a subset of identity traits is considered in the analysis. This limitation is due to the lack of availability of resources specifically designed to account for other identity traits. To overcome this issue, we plan to develop system and resources dedicated to widening the range of identities and their intersection that can be erased by filters and guardrails.  
The manual annotation was performed by two annotators, which is sufficient to establish the directional claim that human evaluators systematically disagree with automated systems, but insufficient to capture the cross-cultural and demographic variation in harm judgements that prior work documents as substantial. We treat the annotation as a first-pass audit calibrated to the scope of this study rather than as a reusable benchmark. 

\section*{Ethical Consideration}
The research involved two interns who annotated texts flagged for content moderation as part of their internship activities. Their work was closely supervised by a senior postdoctoral researcher, who ensured appropriate guidance and support throughout the annotation process. Additionally, we asked the authorization from the Bioethical Committee of one of the authors' institution.

By demonstrating that filters and guardrails can exhibit systematic biases, we aim to foster discussion about the unintended side effects of content moderation systems. Although this is a complex and sensitive issue, we recognize that some actors may misuse these findings to argue against laws and measures designed to combat hate speech and other forms of discrimination. We therefore emphasize that our objective is to develop more effective and equitable approaches to safeguarding vulnerable and marginalized communities.

We adopted an AI assistant for grammatical check during paper writing.

\bibliography{custom}

@inproceedings{dodge-etal-2021-documenting,
    title = "Documenting Large Webtext Corpora: A Case Study on the Colossal Clean Crawled Corpus",
    author = "Dodge, Jesse  and
      Sap, Maarten  and
      Marasovi{\'c}, Ana  and
      Agnew, William  and
      Ilharco, Gabriel  and
      Groeneveld, Dirk  and
      Mitchell, Margaret  and
      Gardner, Matt",
    editor = "Moens, Marie-Francine  and
      Huang, Xuanjing  and
      Specia, Lucia  and
      Yih, Scott Wen-tau",
    booktitle = "Proceedings of the 2021 Conference on Empirical Methods in Natural Language Processing",
    month = nov,
    year = "2021",
    address = "Online and Punta Cana, Dominican Republic",
    publisher = "Association for Computational Linguistics",
    url = "https://aclanthology.org/2021.emnlp-main.98/",
    doi = "10.18653/v1/2021.emnlp-main.98",
    pages = "1286--1305"
}

@inproceedings{longpre-etal-2024-pretrainers,
    title = "A Pretrainer{'}s Guide to Training Data: Measuring the Effects of Data Age, Domain Coverage, Quality, {\&} Toxicity",
    author = "Longpre, Shayne  and
      Yauney, Gregory  and
      Reif, Emily  and
      Lee, Katherine  and
      Roberts, Adam  and
      Zoph, Barret  and
      Zhou, Denny  and
      Wei, Jason  and
      Robinson, Kevin  and
      Mimno, David  and
      Ippolito, Daphne",
    editor = "Duh, Kevin  and
      Gomez, Helena  and
      Bethard, Steven",
    booktitle = "Proceedings of the 2024 Conference of the North American Chapter of the Association for Computational Linguistics: Human Language Technologies (Volume 1: Long Papers)",
    month = jun,
    year = "2024",
    address = "Mexico City, Mexico",
    publisher = "Association for Computational Linguistics",
    url = "https://aclanthology.org/2024.naacl-long.179/",
    doi = "10.18653/v1/2024.naacl-long.179",
    pages = "3245--3276"
}

@inproceedings{soldaini-etal-2024-dolma,
    title = "Dolma: an Open Corpus of Three Trillion Tokens for Language Model Pretraining Research",
    author = "Soldaini, Luca  and
      Kinney, Rodney  and
      Bhagia, Akshita  and
      Schwenk, Dustin  and
      Atkinson, David  and
      Authur, Russell  and
      Bogin, Ben  and
      Chandu, Khyathi  and
      Dumas, Jennifer  and
      Elazar, Yanai  and
      Hofmann, Valentin  and
      Jha, Ananya  and
      Kumar, Sachin  and
      Lucy, Li  and
      Lyu, Xinxi  and
      Lambert, Nathan  and
      Magnusson, Ian  and
      Morrison, Jacob  and
      Muennighoff, Niklas  and
      Naik, Aakanksha  and
      Nam, Crystal  and
      Peters, Matthew  and
      Ravichander, Abhilasha  and
      Richardson, Kyle  and
      Shen, Zejiang  and
      Strubell, Emma  and
      Subramani, Nishant  and
      Tafjord, Oyvind  and
      Walsh, Evan  and
      Zettlemoyer, Luke  and
      Smith, Noah  and
      Hajishirzi, Hannaneh  and
      Beltagy, Iz  and
      Groeneveld, Dirk  and
      Dodge, Jesse  and
      Lo, Kyle",
    editor = "Ku, Lun-Wei  and
      Martins, Andre  and
      Srikumar, Vivek",
    booktitle = "Proceedings of the 62nd Annual Meeting of the Association for Computational Linguistics (Volume 1: Long Papers)",
    month = aug,
    year = "2024",
    address = "Bangkok, Thailand",
    publisher = "Association for Computational Linguistics",
    url = "https://aclanthology.org/2024.acl-long.840/",
    doi = "10.18653/v1/2024.acl-long.840",
    pages = "15725--15788"
}

@misc{penedo2024finewebdatasetsdecantingweb,
      title={The FineWeb Datasets: Decanting the Web for the Finest Text Data at Scale}, 
      author={Guilherme Penedo and Hynek Kydlíček and Loubna Ben allal and Anton Lozhkov and Margaret Mitchell and Colin Raffel and Leandro Von Werra and Thomas Wolf},
      year={2024},
      eprint={2406.17557},
      archivePrefix={arXiv},
      primaryClass={cs.CL},
      url={https://arxiv.org/abs/2406.17557}, 
}

@inproceedings{lucy-etal-2024-aboutme,
    title = "{A}bout{M}e: Using Self-Descriptions in Webpages to Document the Effects of {E}nglish Pretraining Data Filters",
    author = "Lucy, Li  and
      Gururangan, Suchin  and
      Soldaini, Luca  and
      Strubell, Emma  and
      Bamman, David  and
      Klein, Lauren  and
      Dodge, Jesse",
    editor = "Ku, Lun-Wei  and
      Martins, Andre  and
      Srikumar, Vivek",
    booktitle = "Proceedings of the 62nd Annual Meeting of the Association for Computational Linguistics (Volume 1: Long Papers)",
    month = aug,
    year = "2024",
    address = "Bangkok, Thailand",
    publisher = "Association for Computational Linguistics",
    url = "https://aclanthology.org/2024.acl-long.400/",
    doi = "10.18653/v1/2024.acl-long.400",
    pages = "7393--7420"
}

@misc{mendu2025saferpretraininganalyzingfiltering,
      title={Towards Safer Pretraining: Analyzing and Filtering Harmful Content in Webscale datasets for Responsible LLMs}, 
      author={Sai Krishna Mendu and Harish Yenala and Aditi Gulati and Shanu Kumar and Parag Agrawal},
      year={2025},
      eprint={2505.02009},
      archivePrefix={arXiv},
      primaryClass={cs.CL},
      url={https://arxiv.org/abs/2505.02009}, 
}

@misc{inan2023llamaguardllmbasedinputoutput,
      title={Llama Guard: LLM-based Input-Output Safeguard for Human-AI Conversations}, 
      author={Hakan Inan and Kartikeya Upasani and Jianfeng Chi and Rashi Rungta and Krithika Iyer and Yuning Mao and Michael Tontchev and Qing Hu and Brian Fuller and Davide Testuggine and Madian Khabsa},
      year={2023},
      eprint={2312.06674},
      archivePrefix={arXiv},
      primaryClass={cs.CL},
      url={https://arxiv.org/abs/2312.06674}, 
}

@inproceedings{li-etal-2024-salad,
    title = "{SALAD}-Bench: A Hierarchical and Comprehensive Safety Benchmark for Large Language Models",
    author = "Li, Lijun  and
      Dong, Bowen  and
      Wang, Ruohui  and
      Hu, Xuhao  and
      Zuo, Wangmeng  and
      Lin, Dahua  and
      Qiao, Yu  and
      Shao, Jing",
    editor = "Ku, Lun-Wei  and
      Martins, Andre  and
      Srikumar, Vivek",
    booktitle = "Findings of the Association for Computational Linguistics: ACL 2024",
    month = aug,
    year = "2024",
    address = "Bangkok, Thailand",
    publisher = "Association for Computational Linguistics",
    url = "https://aclanthology.org/2024.findings-acl.235/",
    doi = "10.18653/v1/2024.findings-acl.235",
    pages = "3923--3954"
}

@misc{ghosh2024aegisonlineadaptiveai,
      title={AEGIS: Online Adaptive AI Content Safety Moderation with Ensemble of LLM Experts}, 
      author={Shaona Ghosh and Prasoon Varshney and Erick Galinkin and Christopher Parisien},
      year={2024},
      eprint={2404.05993},
      archivePrefix={arXiv},
      primaryClass={cs.LG},
      url={https://arxiv.org/abs/2404.05993}, 
}

@misc{ji2023beavertailsimprovedsafetyalignment,
      title={BeaverTails: Towards Improved Safety Alignment of LLM via a Human-Preference Dataset}, 
      author={Jiaming Ji and Mickel Liu and Juntao Dai and Xuehai Pan and Chi Zhang and Ce Bian and Chi Zhang and Ruiyang Sun and Yizhou Wang and Yaodong Yang},
      year={2023},
      eprint={2307.04657},
      archivePrefix={arXiv},
      primaryClass={cs.CL},
      url={https://arxiv.org/abs/2307.04657}, 
}

@inproceedings{10.1609/aaai.v37i12.26752,
author = {Markov, Todor and Zhang, Chong and Agarwal, Sandhini and Nekoul, Florentine Eloundou and Lee, Theodore and Adler, Steven and Jiang, Angela and Weng, Lilian},
title = {A holistic approach to undesired content detection in the real world},
year = {2023},
isbn = {978-1-57735-880-0},
publisher = {AAAI Press},
url = {https://doi.org/10.1609/aaai.v37i12.26752},
doi = {10.1609/aaai.v37i12.26752},
booktitle = {Proceedings of the Thirty-Seventh AAAI Conference on Artificial Intelligence and Thirty-Fifth Conference on Innovative Applications of Artificial Intelligence and Thirteenth Symposium on Educational Advances in Artificial Intelligence},
articleno = {1683},
numpages = {10},
series = {AAAI'23/IAAI'23/EAAI'23}
}

@misc{davidson2017automatedhatespeechdetection,
      title={Automated Hate Speech Detection and the Problem of Offensive Language}, 
      author={Thomas Davidson and Dana Warmsley and Michael Macy and Ingmar Weber},
      year={2017},
      eprint={1703.04009},
      archivePrefix={arXiv},
      primaryClass={cs.CL},
      url={https://arxiv.org/abs/1703.04009}, 
}

@inproceedings{caselli-etal-2021-hatebert,
    title = "{H}ate{BERT}: Retraining {BERT} for Abusive Language Detection in {E}nglish",
    author = "Caselli, Tommaso  and
      Basile, Valerio  and
      Mitrovi{\'c}, Jelena  and
      Granitzer, Michael",
    editor = "Mostafazadeh Davani, Aida  and
      Kiela, Douwe  and
      Lambert, Mathias  and
      Vidgen, Bertie  and
      Prabhakaran, Vinodkumar  and
      Waseem, Zeerak",
    booktitle = "Proceedings of the 5th Workshop on Online Abuse and Harms (WOAH 2021)",
    month = aug,
    year = "2021",
    address = "Online",
    publisher = "Association for Computational Linguistics",
    url = "https://aclanthology.org/2021.woah-1.3/",
    doi = "10.18653/v1/2021.woah-1.3",
    pages = "17--25"
}

@misc{davani2021hatespeechclassifierslearn,
      title={Hate Speech Classifiers Learn Human-Like Social Stereotypes}, 
      author={Aida Mostafazadeh Davani and Mohammad Atari and Brendan Kennedy and Morteza Dehghani},
      year={2021},
      eprint={2110.14839},
      archivePrefix={arXiv},
      primaryClass={cs.CL},
      url={https://arxiv.org/abs/2110.14839}, 
}

@article{Stranisci_2026,
   title={What Are They Filtering Out? An Experimental Benchmark of Filtering Strategies for Harm Reduction in Pretraining Datasets},
   volume={40},
   ISSN={2159-5399},
   url={http://dx.doi.org/10.1609/aaai.v40i46.41279},
   DOI={10.1609/aaai.v40i46.41279},
   number={46},
   journal={Proceedings of the AAAI Conference on Artificial Intelligence},
   publisher={Association for the Advancement of Artificial Intelligence (AAAI)},
   author={Stranisci, Marco Antonio and Hardmeier, Christian},
   year={2026},
   month=Mar, pages={39303–39313} }

@misc{gao2020pile800gbdatasetdiverse,
      title={The Pile: An 800GB Dataset of Diverse Text for Language Modeling}, 
      author={Leo Gao and Stella Biderman and Sid Black and Laurence Golding and Travis Hoppe and Charles Foster and Jason Phang and Horace He and Anish Thite and Noa Nabeshima and Shawn Presser and Connor Leahy},
      year={2020},
      eprint={2101.00027},
      archivePrefix={arXiv},
      primaryClass={cs.CL},
      url={https://arxiv.org/abs/2101.00027}, 
}

@article{raffel2020exploring,
  title={Exploring the limits of transfer learning with a unified text-to-text transformer},
  author={Raffel, Colin and Shazeer, Noam and Roberts, Adam and Lee, Katherine and Narang, Sharan and Matena, Michael and Zhou, Yanqi and Li, Wei and Liu, Peter J},
  journal={Journal of machine learning research},
  volume={21},
  number={140},
  pages={1--67},
  year={2020}
}

@inproceedings{bender2021dangers,
  title={On the dangers of stochastic parrots: Can language models be too big?},
  author={Bender, Emily M and Gebru, Timnit and McMillan-Major, Angelina and Shmitchell, Shmargaret},
  booktitle={Proceedings of the 2021 ACM conference on fairness, accountability, and transparency},
  pages={610--623},
  year={2021}
}

@inproceedings{blodgett2020language,
  title={Language (Technology) is Power: A Critical Survey of “Bias” in NLP},
  author={Blodgett, Su Lin and Barocas, Solon and Daum{\'e} III, Hal and Wallach, Hanna},
  booktitle={Proceedings of the 58th Annual Meeting of the Association for Computational Linguistics},
  pages={5454--5476},
  year={2020}
}

@inproceedings{jo2020lessons,
  title={Lessons from archives: Strategies for collecting sociocultural data in machine learning},
  author={Jo, Eun Seo and Gebru, Timnit},
  booktitle={Proceedings of the 2020 conference on fairness, accountability, and transparency},
  pages={306--316},
  year={2020}
}

@article{albalak2024survey,
  title={A survey on data selection for language models},
  author={Albalak, Alon and Elazar, Yanai and Xie, Sang Michael and Longpre, Shayne and Lambert, Nathan and Wang, Xinyi and Muennighoff, Niklas and Hou, Bairu and Pan, Liangming and Jeong, Haewon and others},
  journal={arXiv preprint arXiv:2402.16827},
  year={2024}
}

@inproceedings{xu2021detoxifying,
  title={Detoxifying Language Models Risks Marginalizing Minority Voices},
  author={Xu, Albert and Pathak, Eshaan and Wallace, Eric and Gururangan, Suchin and Sap, Maarten and Klein, Dan},
  booktitle={Proceedings of the 2021 Conference of the North American Chapter of the Association for Computational Linguistics: Human Language Technologies},
  pages={2390--2397},
  year={2021}
}

@inproceedings{luccioni2021s,
  title={What’s in the box? an analysis of undesirable content in the Common Crawl corpus},
  author={Luccioni, Alexandra and Viviano, Joseph},
  booktitle={Proceedings of the 59th Annual Meeting of the Association for Computational Linguistics and the 11th International Joint Conference on Natural Language Processing (Volume 2: Short Papers)},
  pages={182--189},
  year={2021}
}

@inproceedings{erxleben2014introducing,
  title={Introducing wikidata to the linked data web},
  author={Erxleben, Fredo and G{\"u}nther, Michael and Kr{\"o}tzsch, Markus and Mendez, Julian and Vrande{\v{c}}i{\'c}, Denny},
  booktitle={The Semantic Web--ISWC 2014: 13th International Semantic Web Conference, Riva del Garda, Italy, October 19-23, 2014. Proceedings, Part I 13},
  pages={50--65},
  year={2014},
  organization={Springer}

}

@misc{sorensen2024roadmappluralisticalignment,
      title={A Roadmap to Pluralistic Alignment}, 
      author={Taylor Sorensen and Jared Moore and Jillian Fisher and Mitchell Gordon and Niloofar Mireshghallah and Christopher Michael Rytting and Andre Ye and Liwei Jiang and Ximing Lu and Nouha Dziri and Tim Althoff and Yejin Choi},
      year={2024},
      eprint={2402.05070},
      archivePrefix={arXiv},
      primaryClass={cs.AI},
      url={https://arxiv.org/abs/2402.05070}, 
}

@inproceedings{draetta2024reclaim,
  title={ReCLAIM project: Exploring Italian slurs reappropriation with large language models},
  author={Draetta, Lia and Ferrando, Chiara and Cuccarini, Marco and James, Liam and Patti, Viviana},
  booktitle={Proceedings of the 10th Italian Conference on Computational Linguistics (CLiC-it 2024)},
  pages={335--342},
  year={2024}
}

@article{vidgen2020directions,
  title={Directions in abusive language training data, a systematic review: Garbage in, garbage out},
  author={Vidgen, Bertie and Derczynski, Leon},
  journal={Plos one},
  volume={15},
  number={12},
  pages={e0243300},
  year={2020},
  publisher={Public Library of Science San Francisco, CA USA}
}

@inproceedings{sap2019risk,
  title={The risk of racial bias in hate speech detection},
  author={Sap, Maarten and Card, Dallas and Gabriel, Saadia and Choi, Yejin and Smith, Noah A},
  booktitle={Proceedings of the 57th annual meeting of the association for computational linguistics},
  pages={1668--1678},
  year={2019}
}

@book{united1982standard,
  title={Standard country or area codes for statistical use},
  author={United Nations. Statistical Office},
  number={49},
  year={1982},
  publisher={UN}
}

@inproceedings{davani2024disentangling,
  title={Disentangling perceptions of offensiveness: Cultural and moral correlates},
  author={Davani, Aida and D{\'\i}az, Mark and Baker, Dylan and Prabhakaran, Vinodkumar},
  booktitle={Proceedings of the 2024 ACM Conference on Fairness, Accountability, and Transparency},
  pages={2007--2021},
  year={2024}
}

@inproceedings{lee2024exploring,
  title={Exploring cross-cultural differences in English hate speech annotations: From dataset construction to analysis},
  author={Lee, Nayeon and Jung, Chani and Myung, Junho and Jin, Jiho and Camacho-Collados, Jose and Kim, Juho and Oh, Alice},
  booktitle={Proceedings of the 2024 Conference of the North American Chapter of the Association for Computational Linguistics: Human Language Technologies (Volume 1: Long Papers)},
  pages={4205--4224},
  year={2024}
}

@inproceedings{kotek-etal-2023-gender,
author = {Kotek, Hadas and Dockum, Rikker and Sun, David},
title = {Gender bias and stereotypes in Large Language Models},
year = {2023},
isbn = {9798400701139},
publisher = {Association for Computing Machinery},
address = {New York, NY, USA},
url = {https://doi.org/10.1145/3582269.3615599},
doi = {10.1145/3582269.3615599},
booktitle = {Proceedings of The ACM Collective Intelligence Conference},
pages = {12–24},
numpages = {13},
location = {Delft, Netherlands},
series = {CI '23}
}

@inproceedings{sheng-etal-2019-woman,
    title = "The Woman Worked as a Babysitter: On Biases in Language Generation",
    author = "Sheng, Emily  and
      Chang, Kai-Wei  and
      Natarajan, Premkumar  and
      Peng, Nanyun",
    editor = "Inui, Kentaro  and
      Jiang, Jing  and
      Ng, Vincent  and
      Wan, Xiaojun",
    booktitle = "Proceedings of the 2019 Conference on Empirical Methods in Natural Language Processing and the 9th International Joint Conference on Natural Language Processing (EMNLP-IJCNLP)",
    month = nov,
    year = "2019",
    address = "Hong Kong, China",
    publisher = "Association for Computational Linguistics",
    url = "https://aclanthology.org/D19-1339/",
    doi = "10.18653/v1/D19-1339",
    pages = "3407--3412"
}

@inproceedings{dorn-etal-2024-harmful,
author = {Dorn, Rebecca and Kezar, Lee and Morstatter, Fred and Lerman, Kristina},
title = {Harmful Speech Detection by Language Models Exhibits Gender-Queer Dialect Bias},
year = {2024},
isbn = {9798400712227},
publisher = {Association for Computing Machinery},
address = {New York, NY, USA},
url = {https://doi.org/10.1145/3689904.3694704},
doi = {10.1145/3689904.3694704},
booktitle = {Proceedings of the 4th ACM Conference on Equity and Access in Algorithms, Mechanisms, and Optimization},
articleno = {6},
numpages = {12},
location = {San Luis Potosi, Mexico},
series = {EAAMO '24}
}

@inproceedings{ovalle-etal-2025-root,
author = {Ovalle, Anaelia and Pavasovic, Krunoslav Lehman and Martin, Louis and Zettlemoyer, Luke and Smith, Eric Michael and Chang, Kai-Wei and Williams, Adina and Sagun, Levent},
title = {The Root Shapes the Fruit: On the Persistence of Gender-Exclusive Harms in Aligned Language Models},
year = {2025},
isbn = {9798400714825},
publisher = {Association for Computing Machinery},
address = {New York, NY, USA},
url = {https://doi.org/10.1145/3715275.3732196},
doi = {10.1145/3715275.3732196},
booktitle = {Proceedings of the 2025 ACM Conference on Fairness, Accountability, and Transparency},
pages = {3094–3105},
numpages = {12},
location = {
},
series = {FAccT '25}
}

@misc{qadri-etal-2025-risks,
      title={Risks of Cultural Erasure in Large Language Models}, 
      author={Rida Qadri and Aida M. Davani and Kevin Robinson and Vinodkumar Prabhakaran},
      year={2025},
      eprint={2501.01056},
      archivePrefix={arXiv},
      primaryClass={cs.CL},
      url={https://arxiv.org/abs/2501.01056}, 
}

@inproceedings{schwobel-etal-2023-geographical,
    title = "Geographical Erasure in Language Generation",
    author = {Schw{\"o}bel, Pola  and
      Golebiowski, Jacek  and
      Donini, Michele  and
      Archambeau, Cedric  and
      Pruthi, Danish},
    editor = "Bouamor, Houda  and
      Pino, Juan  and
      Bali, Kalika",
    booktitle = "Findings of the Association for Computational Linguistics: EMNLP 2023",
    month = dec,
    year = "2023",
    address = "Singapore",
    publisher = "Association for Computational Linguistics",
    url = "https://aclanthology.org/2023.findings-emnlp.823/",
    doi = "10.18653/v1/2023.findings-emnlp.823",
    pages = "12310--12324"
}

@article{mcglynn2025epistemic,
  author  = {McGlynn, Aidan},
  title   = {Epistemic Objectification in Pornography},
  journal = {The Philosophical Quarterly},
  year    = {2025},
  doi     = {10.1093/pq/pqaf032}
}

@article{gomes2024problematizing,
  author  = {Gomes, Andr{\'e} Belchior and Sultan, Aysel},
  title   = {Problematizing Content Moderation by Social Media Platforms and Its Impact on Digital Harm Reduction},
  journal = {Harm Reduction Journal},
  volume  = {21},
  number  = {1},
  pages   = {194},
  year    = {2024},
  doi     = {10.1186/s12954-024-01104-9}
}

@misc{carlini2021extracting,
      title={Extracting Training Data from Large Language Models}, 
      author={Nicholas Carlini and Florian Tramer and Eric Wallace and Matthew Jagielski and Ariel Herbert-Voss and Katherine Lee and Adam Roberts and Tom Brown and Dawn Song and Ulfar Erlingsson and Alina Oprea and Colin Raffel},
      year={2021},
      eprint={2012.07805},
      archivePrefix={arXiv},
      primaryClass={cs.CR},
      url={https://arxiv.org/abs/2012.07805}, 
}

@inproceedings{katzman2023taxonomizing,
author = {Katzman, Jared and Wang, Angelina and Scheuerman, Morgan and Blodgett, Su Lin and Laird, Kristen and Wallach, Hanna and Barocas, Solon},
title = {Taxonomizing and measuring representational harms: a look at image tagging},
year = {2023},
isbn = {978-1-57735-880-0},
publisher = {AAAI Press},
url = {https://doi.org/10.1609/aaai.v37i12.26670},
doi = {10.1609/aaai.v37i12.26670},
booktitle = {Proceedings of the Thirty-Seventh AAAI Conference on Artificial Intelligence and Thirty-Fifth Conference on Innovative Applications of Artificial Intelligence and Thirteenth Symposium on Educational Advances in Artificial Intelligence},
articleno = {1601},
numpages = {9},
series = {AAAI'23/IAAI'23/EAAI'23}
}

@misc{kirk2021bias,
      title={Bias Out-of-the-Box: An Empirical Analysis of Intersectional Occupational Biases in Popular Generative Language Models}, 
      author={Hannah Kirk and Yennie Jun and Haider Iqbal and Elias Benussi and Filippo Volpin and Frederic A. Dreyer and Aleksandar Shtedritski and Yuki M. Asano},
      year={2021},
      eprint={2102.04130},
      archivePrefix={arXiv},
      primaryClass={cs.CL},
      url={https://arxiv.org/abs/2102.04130}, 
}

@misc{rathi2026shaping,
      title={Shaping capabilities with token-level data filtering}, 
      author={Neil Rathi and Alec Radford},
      year={2026},
      eprint={2601.21571},
      archivePrefix={arXiv},
      primaryClass={cs.LG},
      url={https://arxiv.org/abs/2601.21571}, 
}

@inproceedings{feng2023pretraining,
    title = "From Pretraining Data to Language Models to Downstream Tasks: Tracking the Trails of Political Biases Leading to Unfair {NLP} Models",
    author = "Feng, Shangbin  and
      Park, Chan Young  and
      Liu, Yuhan  and
      Tsvetkov, Yulia",
    editor = "Rogers, Anna  and
      Boyd-Graber, Jordan  and
      Okazaki, Naoaki",
    booktitle = "Proceedings of the 61st Annual Meeting of the Association for Computational Linguistics (Volume 1: Long Papers)",
    month = jul,
    year = "2023",
    address = "Toronto, Canada",
    publisher = "Association for Computational Linguistics",
    url = "https://aclanthology.org/2023.acl-long.656/",
    doi = "10.18653/v1/2023.acl-long.656",
    pages = "11737--11762"
}

@inproceedings{queer-in-ai-dni-guide,
   title = "How to Make Virtual Conferences Queer-Friendly: A Guide",
   author = "QueerInAI, Organizers of and Pranav, A and Bleile, MaryLena and Subramonian, Arjun and Soldaini, Luca and Sutherland, Danica J. and Weber, Sabine and Xu, Pan",
   booktitle = "Proceedings of the 2021 Workshop on Widening NLP",
   month = nov,
   year = "2021",
   address = "Punta Cana, Dominican Republic",
   publisher = "Conference on Empirical Methods in Natural Language Processing",
   doi = "queerinai.org/diversity-guide",
}

@book{kukutai2016indigenous,
  title={Indigenous data sovereignty: Toward an agenda},
  author={Kukutai, Tahu and Taylor, John},
  volume={38},
  year={2016},
  publisher={ANU press}
}

@misc{ganguli2022redteaminglanguagemodels,
      title={Red Teaming Language Models to Reduce Harms: Methods, Scaling Behaviors, and Lessons Learned}, 
      author={Deep Ganguli and Liane Lovitt and Jackson Kernion and Amanda Askell and Yuntao Bai and Saurav Kadavath and Ben Mann and Ethan Perez and Nicholas Schiefer and Kamal Ndousse and Andy Jones and Sam Bowman and Anna Chen and Tom Conerly and Nova DasSarma and Dawn Drain and Nelson Elhage and Sheer El-Showk and Stanislav Fort and Zac Hatfield-Dodds and Tom Henighan and Danny Hernandez and Tristan Hume and Josh Jacobson and Scott Johnston and Shauna Kravec and Catherine Olsson and Sam Ringer and Eli Tran-Johnson and Dario Amodei and Tom Brown and Nicholas Joseph and Sam McCandlish and Chris Olah and Jared Kaplan and Jack Clark},
      year={2022},
      eprint={2209.07858},
      archivePrefix={arXiv},
      primaryClass={cs.CL},
      url={https://arxiv.org/abs/2209.07858}, 
}

@misc{khandelwal2020generalizationmemorizationnearestneighbor,
      title={Generalization through Memorization: Nearest Neighbor Language Models}, 
      author={Urvashi Khandelwal and Omer Levy and Dan Jurafsky and Luke Zettlemoyer and Mike Lewis},
      year={2020},
      eprint={1911.00172},
      archivePrefix={arXiv},
      primaryClass={cs.CL},
      url={https://arxiv.org/abs/1911.00172}, 
}
\appendix
\section{Detailed results on study of epistemic erasure of marginalised identities}
\label{app:rates}

Tables~\ref{tab:app_trans}, \ref{tab:app_gender}, and~\ref{tab:app_origin}
report per-system flag rates for every (group, system) cell along the
three identity axes. Each cell is the percentage of sentences in that
group that the named system flagged. Cell shading marks where the rate
ratio against the reference departs from 1.0 enough to carry signal
(red for over-flagging, blue for under-flagging, deeper colour for a
larger departure); cell typography encodes the BH-FDR significance of
that departure (\textbf{bold} $q<.001$, plain $q<.05$, \textit{italic}
not significant). Cells whose absolute rate falls below the 0.5\% noise
floor are left unshaded because their ratios are unstable.

\paragraph{Gender modality dimension.}
Cisgender flag rates establish the baseline. Shutterstock catches
$48.5\%$ of cisgender mentions, Qwen3Guard $43.6\%$, MD-Judge $33.0\%$,
and Llama-Guard $21.5\%$. The remaining three systems sit at lower
baselines: Hatebase $5.7\%$, DOLMA $23.0\%$, HateBERT $0.1\%$. On
transgender mentions, Shutterstock and the three guardrails each move
sharply upward (Shutterstock $76.8\%$, Llama-Guard $38.1\%$, MD-Judge
$48.1\%$, Qwen3Guard $68.3\%$; all $q<.001$ against the cisgender
reference). Hatebase moves in the opposite direction ($3.5\%$,
$q=.012$): its lexicon was tuned on slurs that target other identity
groups, so transgender-coded vocabulary falls outside its match
patterns. DOLMA and HateBERT change by less than three percentage
points each.

\begin{table*}[t]
\centering
\footnotesize
\setlength{\tabcolsep}{4pt}
\renewcommand{\arraystretch}{1.18}
\definecolor{rrUp2}{HTML}{F4A582}
\definecolor{rrUp1}{HTML}{FDDBC7}
\definecolor{rrDn1}{HTML}{D1E5F0}
\definecolor{rrDn2}{HTML}{92C5DE}
\begin{tabular}{l|cccc|ccc}
\toprule
 & \multicolumn{4}{c|}{\textbf{Filters}} & \multicolumn{3}{c}{\textbf{Guardrails}} \\
\cmidrule(lr){2-5}\cmidrule(lr){6-8}
\textbf{Group} & \textbf{Hatebase} & \textbf{Shutter.} & \textbf{DOLMA} & \textbf{HateBERT} & \textbf{Llama-G.} & \textbf{MD-J.} & \textbf{Qwen3G.} \\
\midrule
Cisgender & 5.7 & 48.5 & 23.0 & 0.1 & 21.5 & 33.0 & 43.6 \\
Transgender & \cellcolor{rrDn1}3.5 & \cellcolor{rrUp1}\textbf{76.8} & \textit{20.5} & \textit{0.1} & \cellcolor{rrUp1}\textbf{38.1} & \textbf{48.1} & \cellcolor{rrUp1}\textbf{68.3} \\
\bottomrule
\end{tabular}
\caption{Per-system flag rates (\%) for the trans/cis dimension. Reference: cisgender. \textbf{Cell shading} marks rate ratios that depart from 1.0: red tones for over-flagging (focal $>$ reference; light $1.5\leq RR<2.0$, dark $RR\geq 2.0$); blue tones for under-flagging (focal $<$ reference; light $0.5<RR\leq 0.7$, dark $RR\leq 0.5$); no shading otherwise. \textbf{Cell typography} encodes BH-FDR significance against the reference: \textbf{bold} for $q<.001$, plain for $q<.05$, \textit{italic} for not significant. Cells whose absolute rate is below 0.5\% receive no shading because the ratio is unstable at this floor.}
\label{tab:app_trans}
\end{table*}

\paragraph{Gender identity dimension.}
Among the four filters, only Shutterstock responds at scale to gender
mentions. It flags $47.3\%$ of man mentions and adds 14--20 percentage
points on woman, non-binary, and questioning mentions ($61.4\%$,
$66.9\%$, $62.3\%$; all $q<.001$). The other three filters are quiet:
Hatebase stays at $5$--$6\%$ across all five groups, HateBERT below
$0.3\%$. DOLMA's only gender signal is on questioning mentions, where
it rises to $50.9\%$ from a man baseline of $23.5\%$ ($q<.001$). The
three guardrails are uniformly more sensitive to non-man gender
mentions: Llama-Guard rises from $19.4\%$ on man to $30$--$32\%$ on
woman, non-binary, and questioning; MD-Judge from $31.3\%$ to
$43$--$58\%$; Qwen3Guard from $42.1\%$ to $53$--$67\%$. Genderfluid /
gender non-conforming mentions sit on a small subsample, so the
confidence intervals are wide and only Shutterstock reaches
significance.

\begin{table*}[t]
\centering
\footnotesize
\setlength{\tabcolsep}{4pt}
\renewcommand{\arraystretch}{1.18}
\definecolor{rrUp2}{HTML}{F4A582}
\definecolor{rrUp1}{HTML}{FDDBC7}
\definecolor{rrDn1}{HTML}{D1E5F0}
\definecolor{rrDn2}{HTML}{92C5DE}
\begin{tabular}{l|cccc|ccc}
\toprule
 & \multicolumn{4}{c|}{\textbf{Filters}} & \multicolumn{3}{c}{\textbf{Guardrails}} \\
\cmidrule(lr){2-5}\cmidrule(lr){6-8}
\textbf{Group} & \textbf{Hatebase} & \textbf{Shutter.} & \textbf{DOLMA} & \textbf{HateBERT} & \textbf{Llama-G.} & \textbf{MD-J.} & \textbf{Qwen3G.} \\
\midrule
Man & 6.0 & 47.3 & 23.5 & 0.1 & 19.4 & 31.3 & 42.1 \\
Woman & \textit{6.0} & \textbf{61.4} & \textit{25.5} & \textit{0.2} & \cellcolor{rrUp1}\textbf{30.1} & \textbf{43.3} & \textbf{56.5} \\
Non-binary & \textit{5.4} & \textbf{66.9} & \textit{18.6} & \textit{0.1} & \textit{18.9} & \textit{23.8} & \textit{52.9} \\
Questioning & \textit{4.8} & \textit{62.3} & \cellcolor{rrUp2}\textbf{50.9} & \textit{0.1} & \cellcolor{rrUp1}\textit{32.0} & \cellcolor{rrUp1}\textit{58.0} & \cellcolor{rrUp1}\textit{66.9} \\
GNC & \textit{5.5} & 61.8 & \textit{30.9} & 0.0 & \cellcolor{rrDn2}1.8 & \cellcolor{rrDn1}\textit{21.8} & \textit{43.6} \\
\bottomrule
\end{tabular}
\caption{Per-system flag rates (\%) for the gender dimension. Reference: man. Shading and typography conventions follow Table~\ref{tab:app_trans}.}
\label{tab:app_gender}
\end{table*}

\paragraph{World-region dimension.}
Central American mentions sit at the floor of Hatebase ($0.2\%$) and
DOLMA ($8.1\%$) and at the ceiling of Shutterstock ($99.3\%$),
Llama-Guard ($95.9\%$), MD-Judge ($65.9\%$), and Qwen3Guard
($98.7\%$). Llama-Guard's contrast with its $26.2\%$ pooled-other
baseline is the largest single deviation in the data set. Mentions of
Western Europe, Western Asia, and Northern Europe move in the
opposite direction from the three guardrails: Llama-Guard flags them
at $6.5\%$, $7.5\%$, and $11.5\%$ respectively, around a quarter of
its baseline; MD-Judge and Qwen3Guard show smaller but consistent
declines on the same three regions. Hatebase reverses on Western
Europe ($15.1\%$ vs.\ $7.5\%$ baseline) and Western Asia ($15.8\%$);
the lexicon's slur coverage skews toward content that mentions
non-Western groups, so the rate on Western-mention text registers
above its pooled-other rate. Eastern Asian mentions are flagged at
$82.2\%$ by Shutterstock and $82.7\%$ by Qwen3Guard, while DOLMA and
Llama-Guard treat them at near-baseline rates. Eastern Europe and
Southern Asia each have one significant per-system cell; Northern
America, the Caribbean, and Australia and New Zealand carry no
significant departures from the pooled-other reference.

\begin{table*}[t]
\centering
\footnotesize
\setlength{\tabcolsep}{4pt}
\renewcommand{\arraystretch}{1.18}
\definecolor{rrUp2}{HTML}{F4A582}
\definecolor{rrUp1}{HTML}{FDDBC7}
\definecolor{rrDn1}{HTML}{D1E5F0}
\definecolor{rrDn2}{HTML}{92C5DE}
\begin{tabular}{l|cccc|ccc}
\toprule
 & \multicolumn{4}{c|}{\textbf{Filters}} & \multicolumn{3}{c}{\textbf{Guardrails}} \\
\cmidrule(lr){2-5}\cmidrule(lr){6-8}
\textbf{Group} & \textbf{Hatebase} & \textbf{Shutter.} & \textbf{DOLMA} & \textbf{HateBERT} & \textbf{Llama-G.} & \textbf{MD-J.} & \textbf{Qwen3G.} \\
\midrule
Central America & 0.2 & \cellcolor{rrUp1}\textbf{99.3} & \cellcolor{rrDn2}\textit{8.1} & \textit{0.4} & \cellcolor{rrUp2}\textbf{95.9} & \cellcolor{rrUp1}65.9 & \cellcolor{rrUp1}\textbf{98.7} \\
Western Europe & \cellcolor{rrUp2}15.1 & \textit{46.4} & \textit{29.0} & 0.0 & \cellcolor{rrDn2}\textbf{6.5} & \cellcolor{rrDn1}26.9 & \cellcolor{rrDn1}37.9 \\
Western Asia & \cellcolor{rrUp2}\textit{15.8} & \cellcolor{rrDn1}34.6 & \textit{29.8} & \textit{0.3} & \cellcolor{rrDn2}7.5 & \cellcolor{rrDn1}24.7 & \cellcolor{rrDn1}38.7 \\
Northern Europe & \cellcolor{rrUp1}\textit{9.4} & 49.2 & \textit{32.7} & \textit{0.1} & \cellcolor{rrDn2}11.5 & \cellcolor{rrDn1}\textit{28.9} & \cellcolor{rrDn1}41.3 \\
Eastern Asia & \cellcolor{rrDn2}\textit{3.0} & \textit{82.2} & \textit{31.0} & 0.0 & \cellcolor{rrUp1}\textit{51.8} & 59.9 & 82.7 \\
Eastern Europe & \cellcolor{rrUp1}\textit{12.7} & \textit{76.2} & \cellcolor{rrDn2}\textit{11.1} & 0.0 & \textit{44.4} & \cellcolor{rrDn2}19.0 & \textit{74.6} \\
Southern Asia & \cellcolor{rrDn2}1.8 & \cellcolor{rrDn1}\textit{43.4} & \cellcolor{rrUp1}\textit{45.2} & \cellcolor{rrUp2}\textit{0.9} & \textit{25.8} & \cellcolor{rrDn1}\textit{23.1} & \cellcolor{rrDn1}\textit{32.1} \\
Northern America & \textit{6.7} & \textit{64.7} & \textit{22.4} & \textit{0.2} & \textit{30.3} & \textit{41.3} & \textit{61.5} \\
Caribbean & \cellcolor{rrDn2}\textit{3.3} & \textit{70.9} & \textit{27.8} & 0.0 & \textit{31.7} & \textit{39.5} & \textit{56.2} \\
Australia/NZ & \textit{8.5} & \textit{70.0} & \textit{20.0} & 0.0 & \textit{36.9} & \cellcolor{rrDn1}\textit{26.2} & \textit{46.9} \\
\bottomrule
\end{tabular}
\caption{Per-system flag rates (\%) for the world-region dimension. Reference: pooled non-target regions. Regions are ordered by aggregate significance against the reference (top to bottom). Shading and typography conventions follow Table~\ref{tab:app_trans}.}
\label{tab:app_origin}
\end{table*}

\end{document}